\documentclass{article}
\usepackage[preprint]{neurips_like}
\usepackage{ms_style}
\shorttitle{Self-Designed Evaluators and Warm Memory for Long-Horizon Agents}
\usepackage[T1]{fontenc}
\usepackage[utf8]{inputenc}
\usepackage{booktabs,graphicx,amsmath,amssymb,xcolor,multirow,array}
\usepackage{microtype}
\usepackage{algorithm}
\usepackage{algpseudocode}
\usepackage[hidelinks]{hyperref}
\graphicspath{{figs/}}
\newcommand{\method}{\textsc{SelfSuite}}
\newcommand{\fde}{\textsc{FDE}}
\newcommand{\best}[1]{\textbf{#1}}

\newcommand{\pp}{\,pp}
\newcommand{\ci}[2]{[#1,\,#2]}
\title{Self-Designed Evaluators and Warm Memory\\ for Long-Horizon Agents}
\author{Saeid Asgari, Emre Kiciman, Leonardo de Oliveira Nunes, Ranveer Chandra \\   \ \ \\
\texttt{saeidasgari@microsoft.com}}

\begin{document}
\maketitle

\begin{abstract}
A tool-using language-model agent deployed over a long horizon of tasks receives no reward: it cannot tell whether it succeeded, so it cannot safely retry, and it cannot label the experience it would need to improve. We present \method{}, in which the agent's own base model, given only the world's public materials and no ground truth, designs a small evaluation suite of weighted judges and a grounded per-task brief, freezes it, and deploys it as the controller of a keep-best retry loop and a typed, outcome-tracked memory. On matched five-repeat benchmarks over tau2-bench and AppWorld with GPT-5.5, \method{} scores above the plain agent without using a single label, matches methods given ten expert labels on tau2-bench, and trails Agentic Context Engineering (ACE) on AppWorld, where code execution gives every learner a direct success signal. In an ablation campaign in which label-free baselines ran on the same tasks, \method{} is above label-free ACE in every repeat, and component ablations identify the second attempt gated by the frozen suite as the only component whose removal hurt in every repeat; memory polarity, the task brief and judge decomposition are within noise. Attempt-level analysis explains why: the self-designed suites are conservative, the keep-best rule turns that conservatism into a low-risk retry gain, and a best-attempt oracle gap remains as headroom. Beyond the label-free setting, we ask how much the one human signal a real deployment has would add: the subject-matter expert (SME) or forward-deployed engineer (\fde{}) who onboards the agent and grades a handful of its early tasks. Simulating that expert with ten labelled onboarding tasks per world, we spend the labels in the two ways an engineer could. Calibrating \method{}'s evaluator through a deterministic, evaluator-only procedure with a strict acceptance gate yields a small but consistent gain; seeding the memory of a strong context-engineering baseline (Agentic Context Engineering, ACE) with the same expert-graded experience lifts that baseline to tie calibrated \method{}. A single-run study on a second model family reproduces the ordering.
\end{abstract}

\section{Introduction}
\label{sec:intro}
An agent that serves a long stream of tasks in one world could, in principle, get better at that world as it goes: retry when it has failed, and remember what worked. The bottleneck for such self-improvement is verification rather than action selection. At deployment there is no reward function, no hidden test and no human to say whether a trajectory was correct. Without such a signal the agent cannot decide whether to retry, and it cannot tell a useful lesson from a misleading one, so it cannot improve over a long stream of tasks. In practice the signal is obtained by an expert hand-authoring checks (success rubrics, policy assertions, world-state read-backs, unit tests), which is laborious, world-specific and brittle, and is the reason most ``self-improving agent'' results assume a benchmark that ships ground truth (GT)~\cite{reflexion,expel,ace}.

We find that the model can author its own verifier, and that in our benchmarks this is enough for the agent to improve on its plain baseline. Given the same public materials a human would read, the base model designs a small, decomposed evaluation suite, freezes it, and uses it, unchanged and without any ground truth, to gate a retry loop and to label experience for memory. We call the method \method{} (\S\ref{sec:method}); it is the object of study in every experiment that follows, and we evaluate it under a protocol designed to survive run-to-run noise.

A real deployment does, however, have one human signal: the forward-deployed engineer (\fde{}) or subject-matter expert (SME) who onboards the agent and can realistically grade about ten tasks per world, and nothing afterwards. We therefore also ask how much that expert adds. We simulate the expert with ten labelled onboarding tasks per world and spend them in the two ways an engineer could (\S\ref{sec:fde}): as \emph{evaluator calibration}, an evaluator-only procedure that uses six labels to propose a revised folding, aggregation and threshold for \method{}'s suite and the other four as a strict acceptance gate, or as \emph{warm memory}, expert-graded experience that seeds a playbook learner before deployment, which also gives \method{} its strongest, budget-matched baseline. The answer is that the expert's labels add a little more, in either way.

\textbf{Contributions.} (1) \method{}: a self-designed, frozen, GT-free control suite with a grounded per-task brief, a world-state read-back that judges consume as evidence, a keep-best retry rule, and a typed profit-and-loss (PnL) memory. (2) A matched, five-repeat evaluation of \method{} across two benchmarks and five worlds with a reported noise floor, paired confidence intervals, component ablations, attempt-level diagnostics, and a second-model replication. (3) A simulated-SME study: ten labelled onboarding tasks per world, excluded from every scored set, spent either as a deterministic evaluator-only calibration of \method{} with a strict fit/gate split or as warm memory for the strongest playbook baseline. All learning is GT-free at deployment; benchmark ground truth is used only for offline scoring.

\section{Related work}
\label{sec:related}
\textbf{Self-improving agents and experiential memory.} A large body of work lets a large language model (LLM) agent improve without weight updates by writing experience into an external memory. Reflexion~\cite{reflexion} stores verbal self-reflections between trials; ExpeL~\cite{expel} extracts insights from successful and failed trajectories; Voyager~\cite{voyager} and Agent Workflow Memory~\cite{awm} accumulate reusable skills and workflows; generative agents~\cite{generative_agents}, MemGPT~\cite{memgpt}, MemoryBank~\cite{memorybank}, A-MEM~\cite{a_mem} and Mem0~\cite{mem0} study how memory is organised, consolidated and retrieved; Buffer of Thoughts~\cite{buffer_of_thoughts}, Dynamic Cheatsheet~\cite{dynamic_cheatsheet} and Learn-by-Interact~\cite{learn_by_interact} distil reusable problem-solving templates or synthesised experience; Memento~\cite{memento} and AdaMEM~\cite{adamem} cast memory use as case-based or test-time adaptation. Agentic Context Engineering (ACE)~\cite{ace} evolves a structured playbook with a generator, reflector and curator, and identifies ``context collapse'' under monolithic rewriting and ``brevity bias'' as failure modes; we use its released memory machinery as our strongest baseline. Surveys cover memory mechanisms~\cite{memory_survey_2024,memory_survey_2026} and self-evolving agents more broadly~\cite{self_evolving_survey,self_evolving_survey2}, and Evo-Memory~\cite{evo_memory} benchmarks test-time learning over task streams. The learning signal is the common weak point. Most of these methods read environment success, unit tests or benchmark labels; ReasoningBank~\cite{reasoningbank} replaces them with an LLM judge's self-assessment of each trajectory, and ACE can run without labels by relying on execution feedback and its reflector's own judgement; and Asadolahi et al.~\cite{memory_reward_inflation} show that such self-assessed scores inflate on incorrect episodes and that the inflation compounds through retrieval. \method{} targets exactly this gap: the signal that labels memory and gates retries is a designed, frozen, state-grounded suite, and every lesson carries a ledger of the outcomes that followed its use.

\textbf{Prompt and context optimisation.} Automatic prompt engineering~\cite{ape}, optimisation by prompting (OPRO)~\cite{opro}, Promptbreeder~\cite{promptbreeder}, DSPy~\cite{dspy} and its instruction optimiser~\cite{miprov2}, TextGrad~\cite{textgrad} and GEPA, a genetic-Pareto reflective prompt optimiser~\cite{gepa}, optimise prompts or programs against a metric computed on labelled examples; see Mei et al.~\cite{context_eng_survey} for a survey of context engineering. These methods assume the metric is available during optimisation. At deployment our agent has no metric, so we optimise nothing online: the evaluator is authored once, frozen, and at most recalibrated from ten expert labels before deployment.

\textbf{LLM-as-a-judge and self-evaluation.} LLM judges are now standard for scoring open-ended outputs~\cite{llmjudge,geval,prometheus,judge_survey}, with panels~\cite{poll} and debate~\cite{chateval} used to reduce individual judge bias. Self-feedback loops such as Self-Refine~\cite{selfrefine}, CRITIC~\cite{critic} and Self-Debug~\cite{self_debug}, AI feedback~\cite{constitutional}, self-rewarding models~\cite{self_rewarding}, self-taught evaluators~\cite{self_taught_eval} and generative verifiers~\cite{generative_verifiers} use a model's own judgements to improve its outputs, building on evidence that models are partly calibrated about their own correctness~\cite{kadavath}. The limits are well documented: without external feedback, intrinsic self-correction often fails to improve reasoning~\cite{cannot_self_correct,kamoi}, self-refinement amplifies self-bias~\cite{self_bias}, and LLM evaluators recognise and favour their own generations~\cite{self_preference}. Our judges are the agent's own model, so these concerns apply directly. \method{} mitigates rather than removes them in three ways: judges are decomposed into narrow criteria with a veto rule, they read the resulting world state rather than only the transcript, and they only gate a retry, which keep-best selection adopts only when the suite prefers it by a margin.

\textbf{Rubrics, checklists and co-evolving evaluators.} Decomposing a judgement into explicit criteria improves evaluation, from atomic fact checking~\cite{factscore} to generated checklists~\cite{tick} and calibrated multi-dimensional rubrics~\cite{llm_rubric}. Rubrics and checklists are increasingly used as rewards for training beyond verifiable domains~\cite{checklists_rm,rubrics_as_rewards,rubric_anchors,openrubrics}, with expert-written rubrics as evaluation targets in domains such as health~\cite{healthbench}; see~\cite{rubrics_survey} for a survey. Recent work generates instance-specific rubrics without human annotation~\cite{dynamic_rubrics,genrubric}, adapts rubrics to agent tasks~\cite{adarubric}, and co-evolves the evaluator with the policy during training~\cite{dynamicrubric,arco,evolm} or evolves evaluation metrics alongside agent skills under a small reference benchmark~\cite{who_grades_grader}. These methods produce training rewards for a policy that is being optimised. \method{} instead has the deployed agent's base model author one world-level suite from public materials, including the per-task brief policy, aggregation rule and retry policy, and freeze it for deployment; the model's weights never change. In our study, an online variant that kept revising the rubric from disagreement mining did not improve on the frozen suite (\S\ref{sec:discussion}).

\textbf{Evaluating agent trajectories.} Automatic evaluators for agents range from vision-language evaluators that also drive refinement~\cite{pan_autonomous} and agentic judges that inspect intermediate work~\cite{agent_as_judge} to judges that actively query the environment for evidence~\cite{aj_bench}. Studies of these evaluators find that LLM judges often accept failed web trajectories~\cite{agentrewardbench}, that reported web-agent progress is overestimated~\cite{illusion_progress}, that judges miss silent process faults that an outcome-only view cannot see~\cite{trajectory_judge}, and struggle to localise errors in long traces~\cite{trail}. Closest to our setting, RubricForge~\cite{rubricforge} evolves a frozen judging rubric against trajectories labelled with environment reward on $\tau$-bench and roughly halves false passes, and DeepVerifier~\cite{rubric_verification_agents} uses rubric-guided verification to drive test-time self-refinement of research agents. Both rely on labelled trajectories or a failure taxonomy built offline; \method{} authors its suite without labels, and our ten-label calibration only re-folds and re-thresholds judges that the model already wrote. The same asymmetry, that a false pass ships a wrong answer while a false fail only costs a retry, motivates our conservative veto design.

\textbf{Verification and test-time selection.} Verifiers that rank sampled solutions~\cite{cobbe,lightman}, voting~\cite{self_consistency}, generated tests~\cite{codet} and search over reasoning or action trees~\cite{tot,lats} convert extra samples into accuracy when a good selector exists. Repeated sampling raises coverage far faster than selection can exploit it~\cite{monkeys}, compute-optimal test-time scaling depends on verifier quality~\cite{snell}, multiple aspect verifiers can replace a trained reward model~\cite{mav}, and for agents, knowing when to reflect and how to verify and merge rollouts matters~\cite{tts_agents}. Reinforcement learning from majority votes~\cite{ttrl} or from the model's own confidence~\cite{intuitor} removes external rewards from training altogether. Our second attempt is the smallest instance of test-time scaling, and our best-attempt oracle gap is the agent-level analogue of the coverage-versus-selection gap: the remaining headroom is in the selector, not in generation.

\textbf{Few-label evaluator alignment.} EvalGen~\cite{evalgen} aligns LLM-generated evaluation assertions with a user's grades on a handful of outputs, and AutoCalibrate~\cite{autocalibrate} drafts and refines a judge's scoring criteria from human labels. Our expert calibration follows the same idea under a stricter protocol for deployment: labels are split into fit and gate sets, only the folding, aggregation and threshold of already-authored judges may change, and a proposal is adopted only if it improves agreement on held-out gate labels without increasing false passes or false fails.

\textbf{Benchmarks and evaluation practice.} We build on ReAct~\cite{react} agents and evaluate on $\tau$-bench~\cite{tau} and its dual-control successor $\tau^2$-bench~\cite{tau2} and on AppWorld~\cite{appworld}; related agent benchmarks include AgentBench~\cite{agentbench}, WebArena~\cite{webarena} and SWE-bench~\cite{swebench}. Both of our benchmarks ship ground truth, which we use only for offline scoring. Following calls for error bars and variance reporting in LLM evaluation~\cite{error_bars,eval_variance} and for cost-aware, reproducible agent evaluation~\cite{agents_that_matter,hal}, we report five matched repeats, paired intervals and the run-to-run noise floor.

\section{Problem setting and the ground-truth boundary}
\label{sec:setting}
An agent solves a stream of tasks $x_1,\dots,x_T$ in a world $\mathcal{W}$ that exposes a public summary, a policy document, tool signatures, and a set of read-only queries over its state. Each attempt produces a trajectory $\tau$ and a resulting world state. Benchmark ground truth $y_t\in\{0,1\}$ exists but is available only to the offline scorer.

We distinguish three regimes, and every reported row is labelled with one. \emph{Deployable, GT-free}: the method never reads $y_t$ for any learning or control-flow decision. \emph{Deployable, expert-onboarded}: the method may read $y_t$ for exactly ten reserved onboarding tasks per world, once, before deployment, standing in for the labels an \fde{} or SME would provide; those tasks are excluded from every scored set. \emph{Oracle ceiling}: the method reads $y_t$ on every task; it is reported only to bound what any signal could buy and is not deployable. Memory is reset at the start of every repeat and is cumulative within a repeat.

\section{\method{}: a self-designed, frozen, ground-truth-free control suite}
\label{sec:method}
\begin{figure}[t]
\centering
\includegraphics[width=\linewidth]{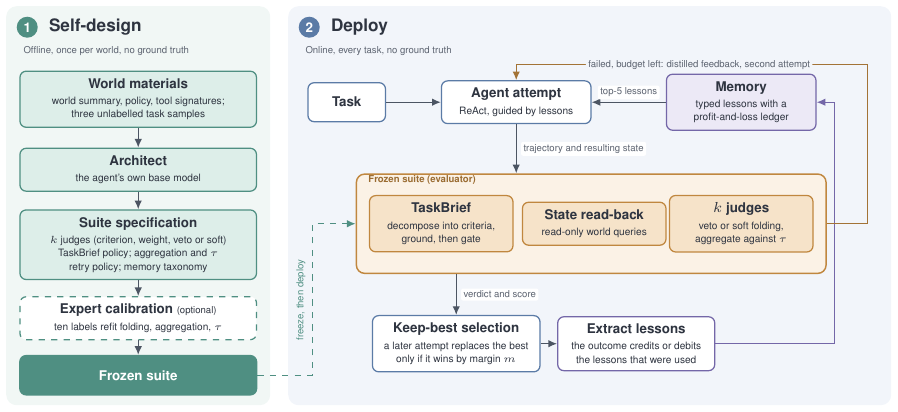}
\caption{\textbf{\method{}.} Left: offline, once per world and without ground truth, the agent's base model (the \emph{architect}) authors a suite specification from the world's public materials; an optional calibration from the expert's ten onboarding labels may revise its folding, aggregation and threshold; the suite is then frozen. Right: online, for every task, the agent attempts the task guided by retrieved lessons; the frozen suite grounds a per-task brief, reads the resulting world state back, and scores the attempt with $k$ judges; a failed attempt with budget left is retried once from distilled feedback; keep-best selection chooses the attempt to submit; lessons are extracted and the outcome updates the memory ledger.}
\label{fig:method}
\end{figure}

\method{} has two phases (Figure~\ref{fig:method}): a one-time \emph{self-design} of the evaluator per world, and an online \emph{deploy} loop in which the frozen evaluator controls retries and labels experience. Both phases are ground-truth-free.

\textbf{Design goals.} The design answers five needs of a real deployment. (i) \emph{No hand-authored checks}: the evaluator is authored from the same public materials an engineer would read, so a new world costs one design call rather than a rubric-writing effort. (ii) \emph{Grounding in the world, not the transcript}: judges read the resulting state back, so ``the transcript looks right'' is not enough to pass. (iii) \emph{Conservative by construction}: veto folding and grounded briefs make false passes rare, which lowers the risk of an automatic retry. (iv) \emph{Outcome-tracked memory}: every lesson carries a ledger, so lessons that precede failures are evicted instead of accumulating. (v) \emph{Frozen and auditable}: the suite does not drift during deployment, its judges can be inspected, and it can be calibrated from a handful of expert labels (\S\ref{sec:fde}).

\subsection{Self-design: the architect authors the suite}
\label{sec:design}
The architect is the agent's own base model. It receives the world summary, the public policy, the tool signatures and, in the \emph{fresh-suite} treatment, three randomly sampled task descriptions without labels, and it emits one JSON suite specification with five parts.

\textbf{Judges.} Each judge is one measurement criterion: a system prompt, a user template over the slots \texttt{\{task, trajectory, policy, tools, lessons, brief, state\}}, an output schema that yields a boolean \texttt{pass}, a \texttt{score} in $[0,1]$ and a \texttt{reason}, a weight, and a \emph{folding} mode. A \textsc{veto} judge is a hard gate: if it fails, the attempt fails. A \textsc{soft} judge contributes its score to a weighted aggregate. Authored suites are small (one to three judges in our campaigns); typical judges are a final-state outcome verifier, a policy or precondition guard, and a side-effect or scope guard (Appendix~\ref{app:suites}).

\textbf{TaskBrief policy.} Judges need to know what ``done'' means for the task at hand. The brief policy decomposes each task instruction into success criteria, required evidence, ordering constraints and constraints, each with a supporting quote from the instruction. Two checks keep the brief honest. \emph{Grounding} asks the model whether each item is supported by the public materials and drops any that is not, so judges cannot invent or relax requirements. \emph{Gating} is a narrow precondition check (did each required action occur at all?) that is conservative by construction: it never fabricates a false negative. The grounded brief conditions every judge through the \texttt{\{brief\}} slot.

\textbf{Aggregation and threshold.} The suite names an aggregation rule (\texttt{min}, \texttt{mean} or \texttt{weighted}) and a pass threshold $\tau$. An attempt passes if no veto judge fails and the aggregate of the soft scores is at least $\tau$.

\textbf{Memory taxonomy and retry policy.} The architect names a small set of task families used to route lessons, and a retry policy (number of attempts and how feedback is formed).

The engine supports an optional self-critique pass over the specification; every configuration in this paper freezes the initial authored suite directly. Judges are independent temperature-0 calls to the agent's own base model; there is no separate, stronger grader.

\subsection{State read-back: evidence, not verdict}
\label{sec:state}
Transcripts are easy to grade leniently; world state is not. After every attempt, the world adapter reconstructs the resulting environment (tau2 replays the final state of the simulation; AppWorld reopens the saved task databases) and runs bounded read-only queries over the records the task concerns, serialising the observations into the \texttt{\{state\}} slot. The read-back never touches benchmark reward, gold actions or hidden tests, and it cannot pass, fail or veto anything by itself: state-conditioned judges interpret it relative to the task and the brief. If the read-back fails, state-conditioned judges abstain neutrally (score $0.5$, flagged \texttt{no\_state}) so a missing observation never creates a false veto. All reported suites condition every judge on the read-back; runtime abstentions were $0\%$ on AppWorld normal and below $1\%$ on AppWorld challenge.

\subsection{Deploy: retrieve, attempt, judge, retry with keep-best, learn}
\label{sec:deploy}
For each task the solver runs the following steps.
\begin{enumerate}
\item \textbf{Retrieve.} The task is routed to a family and the top-5 relevant lessons are retrieved: family-first gather, presort by anchor status and recency, then an LLM relevance rank over a window of about twenty candidates.
\item \textbf{Attempt.} The ReAct agent runs with the retrieved lessons in its prompt.
\item \textbf{Judge.} The frozen suite scores the attempt: the brief is built and grounded, the state is read back, each judge emits \texttt{pass} and \texttt{score}, and the verdicts are folded (vetoes first, then the aggregate against $\tau$).
\item \textbf{Retry.} If the attempt failed and budget remains (two attempts in total), the failing judges' reasons are distilled into one feedback instruction and the agent attempts the task again with that instruction. The feedback describes what the world evidence showed; it never mentions judges or criteria.
\item \textbf{Keep-best.} The attempt to submit is chosen by a do-no-harm rule: a suite-passing attempt wins; ties go to the earliest attempt; a later attempt replaces the current best only if its aggregate score exceeds the best by a margin $m$ ($m=0.30$ in the repeated campaigns, $0.15$ in the single-run study). A noisy second attempt therefore cannot displace a good first one.
\item \textbf{Learn.} Lessons are extracted from the completed task and written to memory, and the outcome updates the ledger of every lesson that was retrieved for the task.
\end{enumerate}

\subsection{Typed, outcome-tracked memory}
\label{sec:memory}
Lessons are typed (positive ``do'' or negative ``avoid''), sectioned (strategies, common mistakes, domain knowledge), routed to the architect's families, and stored in per-bucket-capped lists: at most 30 entries per bucket, LLM distillation of near-duplicates when a bucket reaches 20, and a hard global bound of 200 entries per scope. Every entry carries a PnL ledger with \texttt{uses}, \texttt{helps} and \texttt{hurts} counters: a lesson retrieved before a success is credited and before a failure debited, so persistently harmful lessons are down-weighted and become eligible for eviction. Only five relevance-ranked entries are injected per task, in contrast to a playbook injected as a block. The main treatment uses \emph{balanced} memory (positive and negative lessons); polarity variants are ablated in \S\ref{sec:ablation}.

\section{Simulating the forward-deployed engineer: ten labels, two ways to spend them}
\label{sec:fde}
\method{} needs no labels. A real deployment nevertheless has one human signal, and it is natural to ask what it adds: the agent is onboarded by a forward-deployed engineer or a subject-matter expert who knows the world's policy, watches the agent on a few tasks, and says which ones were done right. We simulate that expert with ten labelled onboarding tasks per world: the agent attempts them once, the expert's verdict on each attempt is the benchmark label, and those ten tasks are then excluded from every scored set. Ten is a budget an engineer can actually deliver during onboarding; it is far below what any training procedure needs. The expert's experience can be spent in two ways.

\subsection{Calibrating the self-designed evaluator}
The calibration (Algorithm~\ref{alg:cal}) is deterministic and \emph{evaluator-only}: it may change which authored judges are kept, their folding, the aggregation rule and the threshold, and nothing else. The actor, the pre-action briefing setting (off), the brief and its gate, the memory and retry policies, the selection margin and the state signal are held fixed, and the ten warm-up tasks are excluded from every scored set.

\begin{algorithm}[t]
\caption{Warm-up-v2 evaluator calibration (six fit tasks, four gate tasks)}
\label{alg:cal}
\begin{algorithmic}[1]
\Require authored suite $S_0$; ten labelled warm-up tasks with recorded per-judge verdicts on fixed attempts
\State Split the ten task identifiers by a seeded shuffle into $F$ (six, fit) and $G$ (four, gate)
\State On $F$, compute each judge's fire rate given correct and given wrong attempts; a judge is \emph{eligible} if $\Pr[\text{fire}\mid\text{wrong}]-\Pr[\text{fire}\mid\text{correct}]\ge0$
\State Enumerate candidates: eligible subset $\times$ folding profile $\in\{$preserve, state-veto, all-soft$\}$ $\times$ aggregation $\in\{$min, mean, weighted$\}$ $\times$ threshold grid (for \texttt{min}, only $\tau_0$)
\State Keep candidates whose agreement and balanced accuracy on $F$ do not fall below $S_0$; propose the candidate maximising (balanced accuracy, agreement, $-$false positives, $-$false negatives, $-$\#judges, $-|\tau-\tau_0|$)
\State \textbf{Accept} the proposal only if, on $G$, agreement strictly improves and neither false positives nor false negatives increase (and the $F$ metrics did not decrease); \textbf{otherwise keep} $S_0$
\State Freeze the resulting suite for the entire deployment
\end{algorithmic}
\end{algorithm}

The calibration operates on \emph{recorded} verdicts: the authored suite is run once on the ten warm-up tasks, every judge's verdict is stored, and candidate suites are re-aggregated from those verdicts without re-running the agent or the judges. The eligibility rule encodes the principle that a veto must be discriminative: a judge that fires on correct attempts as often as on wrong ones only adds retries. With four gate tasks the acceptance rule is deliberately conservative, and it has two consequences. First, the judges only gate retries, so a better evaluator cannot move ground-truth accuracy unless the agent produces a better candidate to select. Second, in the main campaign only the retail proposal cleared the gate (gate agreement $66.7\%\to83.3\%$, a policy-guard judge dropped); the airline proposal would have reduced gate agreement and was rejected; telecom and both AppWorld suites already agreed with all four gate labels, so their authored suites were kept (Appendix~\ref{app:cal}).

\subsection{Seeding memory with expert-graded experience}
The other way to spend the expert's labels is as \emph{warm memory}: the agent's onboarding attempts, together with the expert's verdict on each, are reflected into the agent's playbook before deployment, so the agent starts its label-free stream with lessons that were graded by someone who knew the answer. We implement this with ACE~\cite{ace}: a warm playbook is fitted with ACE's Reflector and Curator from one binary outcome label per onboarding task, after which deployment continues label-free with ACE's own trace self-inference. (Grading ACE's deployment trajectories with the calibrated suite instead of its own trace self-inference is a further arm, reported in Appendix~\ref{app:grid}.)

\section{Experimental setup}
\label{sec:setup}
\textbf{Benchmarks and metrics.} tau2-bench~\cite{tau2} airline (50 tasks), retail (114) and telecom (114): conversational tool-calling agents with a simulated user, scored by pass@0.5 (best-trial reward $\ge0.5$). AppWorld~\cite{appworld} test-normal (168) and test-challenge (417): code-generating agents over app APIs, scored by Task Goal Completion (all subtasks pass, reward $\ge0.999$). A task whose run directory has no evaluation report counts as a failure.

\textbf{Models.} All repeated campaigns use GPT-5.5 at low reasoning effort. The single-run cross-model study adds Claude Opus-4.8 with thinking off. Models are reached through an internal gateway; the judge model is always the agent's own model.

\textbf{Baselines.} A plain ReAct~\cite{react} \textbf{agent}. \textbf{ACE}, driven around our rollouts with the authors' released playbook utilities, the released no-GT Reflector and Curator prompts and delta (accumulating) memory, under a strict no-GT rule: the released code's \texttt{no\_ground\_truth} only hides the target from the Reflector prompt but still gates control flow on correctness, whereas ours never reads benchmark reward for any learning or control decision. \textbf{ACE + warm playbook} (ten labels). In the single-run study the ACE reference is that campaign's own strict no-GT implementation.

\textbf{Campaigns.}
(A) \emph{Matched five-repeat benchmark}: five repeats of 25 tau2 tasks (8 airline; 8 or 9 retail; 9 or 8 telecom) plus 25 AppWorld tasks (12 or 13 from normal and 13 or 12 from challenge, alternating); task identifiers are identical across every method, disjoint across repeats, and disjoint from the ten reserved onboarding tasks per world (airline's 40 evaluation plus 10 onboarding tasks exhaust its catalog). The AppWorld group therefore mixes the two splits; the campaign records the group totals.
(B) \emph{Ablation campaign}: five matched repeats with the same \method{} configuration as (A) (balanced memory, two attempts, margin $0.30$) as reference, seven single-component variants, and the plain agent and ACE without labels (the authors' implementation on AppWorld, our re-implementation on tau2) on the same tasks. Each repeat has 18 or 19 public tau2 tasks (6 or 7 per domain) and 25 AppWorld tasks (12 or 13 normal, 13 or 12 challenge); the campaign also scored a non-public experimental tau2 domain, which we exclude. Per-stratum results separate AppWorld normal from challenge.
(C) \emph{Two-model single-run study}: one run per cell on the full suites for GPT-5.5 (five worlds) and Opus-4.8 (four worlds; its AppWorld-challenge run was invalidated by gateway failures).
(D) \emph{Suite diagnostics}: attempt-level records from the single GPT-5.5 fresh-suite runs of (C) and from five independent deployments of the calibrated suites on every catalog task of airline, retail, telecom and AppWorld normal (the ten onboarding tasks excluded), used in \S\ref{sec:attempts} to characterise how the suites behave rather than to rank methods.

\textbf{Hyperparameters.} Fresh suites are designed from three unlabelled sampled task descriptions; two attempts per task; selection margin $0.30$; balanced memory; briefing off; revision off; state signal required; top-5 retrieval; judge temperature 0; distinct deterministic seeds per repeat and world.

\textbf{Statistics.} We report mean and sample standard deviation (SD) over repeats, paired per-repeat differences with a 95\% $t$ confidence interval (CI) and a paired $t$-test in the matched campaign, Welch's $t$-test for unpaired comparisons, following recommendations to report uncertainty in LLM evaluation~\cite{error_bars,eval_variance}. The observed SD of a 50-task cell is 2 to 7\pp{} and of a 25-task tau2 slice 4 to 13\pp{}; every comparison below should be read against that floor.

\section{Results}
\label{sec:results}

\subsection{Matched five-repeat benchmark}
\label{sec:main}
\begin{figure}[t]
\centering
\includegraphics[width=\linewidth]{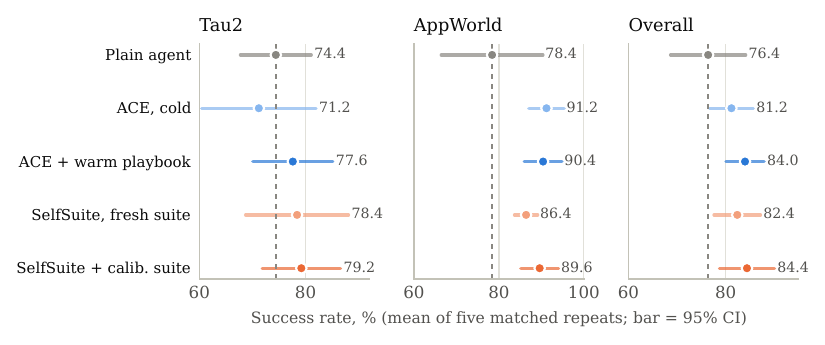}
\caption{\textbf{Matched five-repeat benchmark (campaign A), GPT-5.5.} Mean success over five matched repeats with 95\% CI; the dashed line marks the plain agent. Methods separate on tau2 (policy compliance with no execution signal); on AppWorld the two ACE rows and calibrated \method{} lie within noise of each other, and label-free \method{} is below both ACE rows.}
\label{fig:main}
\end{figure}

\begin{table}[t]
\centering
\caption{\textbf{Paired per-repeat differences (campaign A).} Mean $\pm$ SD of the five per-repeat differences in percentage points, 95\% $t$ CI, and paired $t$-test $p$-value.}
\label{tab:paired}
\scriptsize
\begin{tabular}{l l r r}
\toprule
Comparison & Group & $\Delta$ (pp) & 95\% CI, $p$ \\
\midrule
\method{} + calibrated $-$ plain agent & overall & $+8.0\pm4.7$ & $\ci{+2.2}{+13.8}$, $0.019$ \\
 & tau2 & $+4.8\pm4.4$ & $\ci{-0.6}{+10.2}$, $0.070$ \\
 & AppWorld & $+11.2\pm7.7$ & $\ci{+1.6}{+20.8}$, $0.031$ \\
\method{}, fresh $-$ plain agent & overall & $+6.0\pm5.1$ & $\ci{-0.3}{+12.3}$, $0.058$ \\
ACE + warm playbook $-$ plain agent & overall & $+7.6\pm4.3$ & $\ci{+2.2}{+13.0}$, $0.017$ \\
ACE, cold $-$ plain agent & overall & $+4.8\pm8.6$ & $\ci{-5.8}{+15.4}$, $0.278$ \\
\midrule
\method{} + calibrated $-$ \method{}, fresh & overall & $+2.0\pm1.4$ & $\ci{+0.2}{+3.8}$, $0.034$ \\
 & tau2 & $+0.8\pm3.4$ & $\ci{-3.4}{+5.0}$, $0.621$ \\
 & AppWorld & $+3.2\pm4.4$ & $\ci{-2.2}{+8.6}$, $0.178$ \\
\method{} + calibrated $-$ ACE + warm playbook & overall & $+0.4\pm2.6$ & $\ci{-2.8}{+3.6}$, $0.749$ \\
 & tau2 & $+1.6\pm5.4$ & $\ci{-5.1}{+8.3}$, $0.541$ \\
 & AppWorld & $-0.8\pm3.4$ & $\ci{-5.0}{+3.4}$, $0.621$ \\
\midrule
ACE + warm playbook $-$ ACE, cold & overall & $+2.8\pm5.9$ & $\ci{-4.6}{+10.2}$, $0.351$ \\
 & tau2 & $+6.4\pm9.2$ & $\ci{-5.0}{+17.8}$, $0.195$ \\
\bottomrule
\end{tabular}
\end{table}

Figure~\ref{fig:main} gives the headline comparison (the mean and SD of every arm is listed in Table~\ref{tab:allarms}, Appendix~\ref{app:grid}); Table~\ref{tab:paired} gives the paired differences. Four observations. First, \method{} with a fresh suite and no labels is above the plain agent on average ($+6.0$\pp{} paired, $[-0.3,+12.3]$) and within noise of every learner on tau2, where it has the second-best mean of any method; on AppWorld it is below both ACE rows (paired $-4.8$ and $-4.0$\pp{}, $p=0.03$ each). Second, calibrated \method{} has the best mean overall and on tau2 ($84.4\%$ and $79.2\%$) and ties ACE seeded with expert-graded warm memory, its strongest budget-matched baseline ($84.0\%$; paired difference $+0.4\pm2.6$\pp). Third, the expert's labels help in both uses: as warm memory they lift ACE from 81.2 to 84.0 ($+2.8\pm5.9$\pp{} overall and $+6.4\pm9.2$ on tau2, not significant with five repeats), and as evaluator calibration they are worth a paired $+2.0$\pp{} to \method{} ($[+0.2,+3.8]$), the smallest and most consistent difference in the campaign (per-repeat differences $+2,+2,+2,0,+4$; the calibrated suite never loses a repeat). Fourth, the benchmarks behave differently: on tau2, where policy compliance has no execution signal, the two \method{} rows lead (78.4 and 79.2 versus 71.2 to 77.6) and the methods spread over 8\pp; on AppWorld, where code execution hands every learner a free success signal, both ACE rows and calibrated \method{} lie between 89.6 and 91.2 and within noise of each other, while the fresh suite is the lowest learner (86.4, $5.6$\pp{} below the best, Welch $p=0.06$).

\subsection{Which components matter}
\label{sec:ablation}
\begin{figure}[t]
\centering
\includegraphics[width=\linewidth]{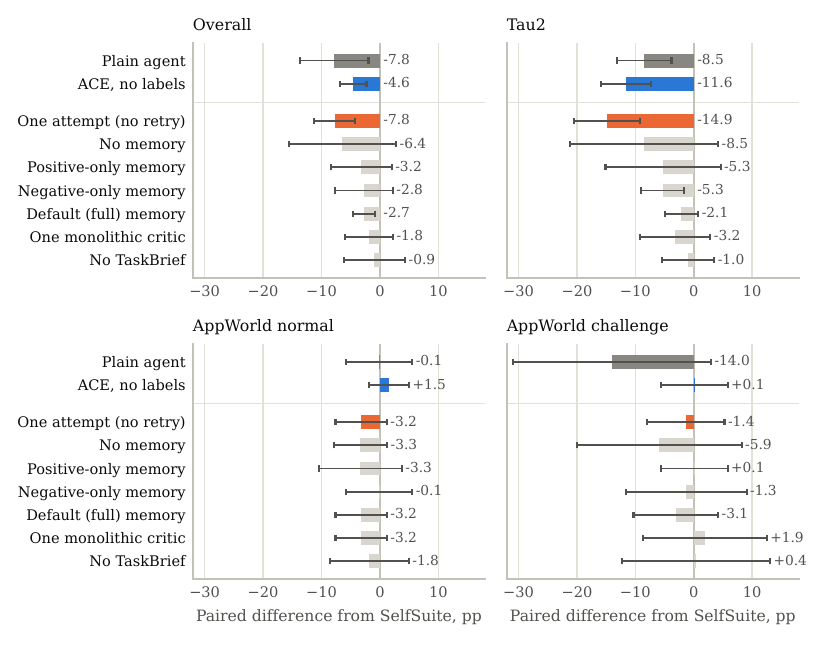}
\caption{\textbf{Component ablations and label-free baselines (campaign B, GPT-5.5).} Paired per-repeat difference from \method{} (balanced memory, two attempts) over five matched repeats of 18 or 19 tau2 and 25 AppWorld tasks (43 or 44 overall); bars are means, whiskers the SD of the five differences. Top: the plain agent and ACE without labels on the same tasks. Bottom: \method{} with one component removed. \method{} scores: overall $89.5\pm5.4$, tau2 $86.2\pm7.0$, AppWorld normal $95.4\pm6.9$, AppWorld challenge $88.6\pm11.1$.}
\label{fig:ablation}
\end{figure}

Figure~\ref{fig:ablation} removes one component at a time from \method{} and, on the same tasks, runs the two label-free baselines. \emph{Baselines.} \method{} scores $89.5\pm5.4\%$ overall against $85.0\pm4.1\%$ for ACE without labels and $81.7\pm4.6\%$ for the plain agent: paired $+4.6\pm2.3$\pp{} over ACE ($\ci{+1.7}{+7.4}$, $p=0.011$, positive in all five repeats) and $+7.8\pm5.8$\pp{} over the agent ($\ci{+0.6}{+15.1}$, $p=0.040$). The margin comes from tau2 ($86.2$ versus $74.6$ and $77.7$); on AppWorld the differences are within noise. \emph{Components.} Removing the second attempt gated by the frozen suite costs $7.8\pm3.5$\pp{} overall ($\ci{-12.1}{-3.5}$, $p=0.007$) and $14.9\pm5.7$\pp{} on tau2, and is negative in all five repeats in both; on AppWorld its effect is small in both splits ($-3.2$ normal, $-1.4$ challenge), consistent with the execution signal that world already provides. Replacing balanced memory by the default memory setting costs a small but consistent $2.7\pm1.9$\pp{} ($p=0.032$, never positive), which supports the balanced design. Removing memory costs $6.4$\pp{} on average but with an SD of $9.2$ (it helps in two repeats and hurts badly in three). Memory polarity, the TaskBrief and judge decomposition (a single monolithic critic without a brief) are within one SD of zero overall, although negative-only memory is below the reference on tau2 in four of five repeats ($-5.3\pm3.7$\pp{}, $p=0.033$).

\subsection{What the suites do: strictness, retry gain and headroom}
\label{sec:attempts}
\begin{figure}[t]
\centering
\includegraphics[width=\linewidth]{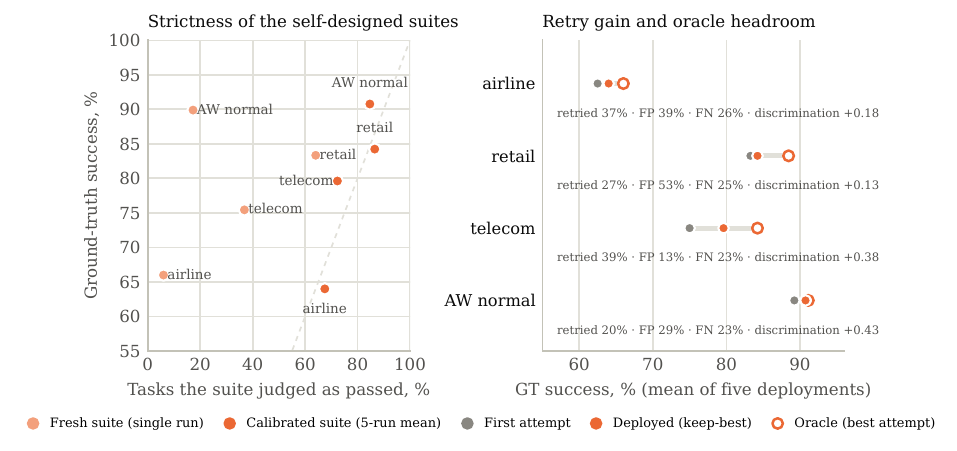}
\caption{\textbf{Attempt-level diagnostics.} Left: fraction of tasks each suite judged as passed versus ground-truth success. Fresh suites (single runs, GPT-5.5) lie far below the diagonal: they are strict vetoers that rarely false-pass. Calibrated suites (five deployments on the whole task sets, means) sit near the diagonal. Right: for the calibrated suite, first-attempt success, deployed keep-best success and the best-attempt oracle, annotated with the fraction of tasks retried and the per-attempt false-positive (FP) and false-negative (FN) rates and discrimination (mean judge score on GT-pass minus GT-fail attempts) of the suite verdict. Judges, aggregation and thresholds of these suites are listed in Appendix~\ref{app:suites}.}
\label{fig:attempts}
\end{figure}

Figure~\ref{fig:attempts} explains the ablation result using the attempt-level records of campaign D (exact values in Table~\ref{tab:operating}, Appendix~\ref{app:numbers}). \emph{Fresh} suites are strict: on single GPT-5.5 runs they judged only $6\%$ (airline), $17\%$ (AppWorld normal), $37\%$ (telecom) and $64\%$ (retail) of tasks as passed while ground-truth success was $66$ to $90\%$, with only 0 to 10 false-pass attempts per world against dozens to hundreds of false fails. Such a suite retries most tasks ($56$ to $100\%$), and because keep-best never lets a second attempt displace a first unless it scores strictly higher by the margin, the retries convert into net gains without net losses: first-attempt success rose from $62.0$ to $66.0$ (airline), $81.6$ to $83.3$ (retail) and $70.2$ to $76.3$ (telecom, full memory), with 3 to 9 tasks gained and 1 to 3 lost per world (AppWorld normal: $90.5\to89.9$, one task lost). \emph{Calibrated} suites are more permissive (judged pass rates $67$ to $87\%$, FN rates $23$ to $26\%$, FP rates $13$ to $53\%$) and retry only $20$ to $39\%$ of tasks, yet they gain $+1.0$ to $+4.6$\pp{} per world and lose on average less than one task per run. The largest gain, telecom ($+4.6$\pp), is also the world with the lowest FP rate and the highest discrimination, which is the pattern one expects if the gain is produced by a vetoer that is right about failures.

The remaining gap to the best-attempt oracle ($+2.0$ airline, $+4.3$ retail, $+4.6$ telecom, $+0.3$ AppWorld normal for the calibrated suite; $+2.4$ to $+12$ on the fresh single runs) is selection headroom: the agent already produces a correct attempt that often, and the GT-free suite does not pick it, the agent-level counterpart of the gap between coverage and selection in repeated sampling~\cite{monkeys}. Better selection would convert directly into task success, which makes the verifier a natural next target.

\subsection{Second model family (single runs)}
\label{sec:crossmodel}
\begin{figure}[t]
\centering
\includegraphics[width=\linewidth]{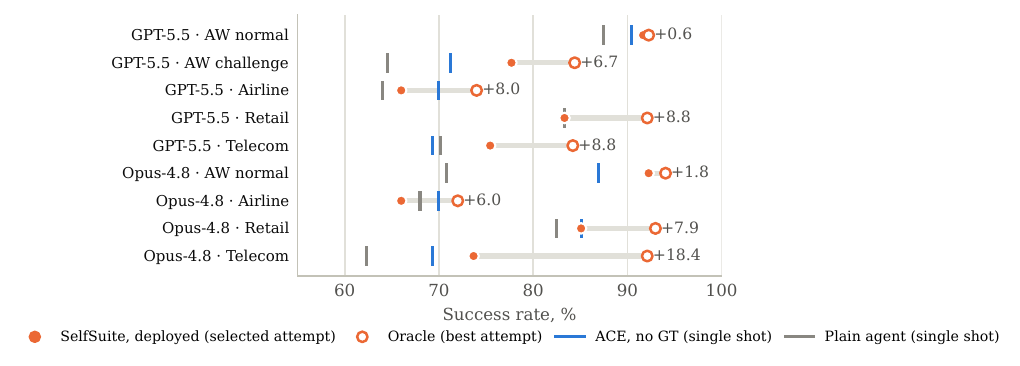}
\caption{\textbf{Two-model single-run study and selection headroom (campaign C).} One run per cell on the full suites. Deployed \method{} success (single-run configuration: no briefing, no revision, selection margin $0.15$), its best-attempt oracle, ACE (that campaign's strict no-GT implementation) and the plain agent. Opus-4.8 AppWorld challenge is omitted (invalid run). The oracle exceeds ACE in all nine cells and the deployed score by $7.4$\pp{} on average.}
\label{fig:headroom}
\end{figure}

On a second model family the ordering reproduces (Figure~\ref{fig:headroom}; exact values in Table~\ref{tab:twomodel}, Appendix~\ref{app:numbers}): \method{} is best or tied on seven of nine cells, ties retail, and loses airline on both models, with lifts over the agent of $+4.9$\pp{} (GPT-5.5) and $+8.4$\pp{} (Opus-4.8) against $+3.0$ and $+6.9$ for that campaign's ACE. Because these are single runs they are supporting evidence for the direction of the effects, not for their size. The best-attempt oracle (Figure~\ref{fig:headroom}) exceeds the deployed score by $7.4$\pp{} on average and by $18.4$\pp{} on Opus-4.8 telecom, and judge discrimination is positive on eight of nine cells (GPT-5.5: $+0.05$, $+0.09$, $+0.09$, $+0.08$, $+0.29$ for AppWorld normal, AppWorld challenge, airline, retail and telecom; Opus-4.8: $+0.16$, $+0.07$, $+0.25$, and $-0.01$ on telecom, the cell with the largest headroom).

\section{Discussion and limitations}
\label{sec:discussion}
\textbf{What \method{} buys, and how.} Without any labels, a frozen self-designed suite with one gated retry puts the agent ahead of the plain agent on average and level with every learner on tau2, the setting with no execution signal, where its mean is above every ACE variant (within noise); on AppWorld, where code execution gives ACE a direct signal, it trails ACE. The ablations and the attempt-level analysis agree on the mechanism: the suites are conservative vetoers, keep-best limits the downside of the retry, and the second attempt is the component whose removal hurts in every repeat. In the ablation campaign, where both label-free baselines ran on the same tasks, \method{} is above ACE without labels in all five repeats and above the plain agent on average. The single-run study on two model families gives the same ordering on every world but airline. What remains on the table is selection: the agent produces a correct attempt more often than the suite picks it.

\textbf{What the expert adds.} In the matched campaign only the plain agent is significantly below the best. Calibrated \method{} has the best mean on tau2 and overall, and ACE seeded with expert-graded warm memory, the strongest baseline once it receives the same ten labels, ties it. The two reach the same operating point by different routes: \method{} spends its effort on a second attempt gated by a conservative self-authored judge; the warm-memory learner spends the expert's labels on a better first attempt.

\textbf{What the expert's ten labels buy.} Spent as warm memory, the labels buy the larger point gain ($+2.8$\pp{} overall and $+6.4$\pp{} on tau2 for the playbook learner, not significant with five repeats): experience that was graded by someone who knew the answer is worth more than the same experience self-graded. Spent as evaluator calibration, they buy a small, consistent gain for \method{} ($+2.0$\pp{} paired, non-negative in every repeat) and a less variable evaluator; the calibration keeps the authored suite in three of five worlds, so the value of the labels is as much a guarantee as an improvement. The two routes are complementary in principle: nothing prevents seeding \method{}'s memory from the same onboarding attempts, which we leave to future work. Grading ACE's trajectories with the calibrated suite instead of its own self-inference adds nothing measurable on top of the warm playbook, and on our context-collapsing re-implementation of ACE the evaluator's effect was to stop drift: a warm playbook that kept adapting label-free lost $3.2$\pp{} on tau2 relative to the same playbook frozen, while adapting under the calibrated evaluator held it (Appendix~\ref{app:grid}).

\textbf{Why retry, and why it is low-risk.} The self-designed suites are conservative vetoers. The keep-best rule limits the downside of retrying: a retry can only replace the first attempt when the suite is more confident in it, so worlds with a discriminative veto (telecom) gain the most and no world loses more than one task per run on average. The same property explains why the judges cannot buy accuracy by themselves: they gate retries; they do not generate candidates.

\textbf{The signal the world already gives.} On AppWorld the ACE rows and calibrated \method{} tie because code execution exposes success to all of them, and label-free \method{} trails ACE; the ablations show the retry effect shrinking to a few points on both AppWorld splits. Claims about GT-free learning are most informative on worlds like tau2, where compliance with a policy leaves no execution trace; that is also where \method{} has its highest means, though within noise of warm ACE.

\textbf{Airline.} \method{} does not beat the plain agent on airline on either model in the single-run study and trails ACE there, and the calibration could not improve the airline suite on the gate tasks. Airline's policy is dense enough that a slightly wrong lesson or judge costs more than it saves. We report it as a limit of the approach rather than tune around it.

\textbf{Limitations.} The judge, the brief grounder, the memory ranker and the agent are the same model; decomposition, grounding and state read-back reduce but do not remove self-evaluation circularity~\cite{self_preference,cannot_self_correct}, and Opus-4.8 telecom shows the residual (negative discrimination with $18$\pp{} of headroom). Fresh suites differ from repeat to repeat, so part of the reported variance is design variance; the calibrated suite removes it within a campaign at the price of freezing whatever the ten warm-up tasks supported. Five repeats bound effects of about $\pm4$\pp{} on a 50-task cell; smaller effects (memory polarity, fresh versus calibrated suites on tau2) are not resolved by this study. AppWorld challenge appears in the single-run study and in the ablation campaign but not on its own in the matched campaign, and Opus-4.8 only in the single-run study. The comparison is not cost-normalised: \method{} adds judge, brief, retrieval and retry calls per task, and a per-task token accounting, as advocated for agent evaluation~\cite{agents_that_matter,hal}, is future work. Finally, an online variant in which the rubric keeps being revised from GT-free disagreement mining did not improve on the frozen calibrated suite in the matched campaign; whether windowed rubric evolution can help with a larger adjudication budget is open.

\section{Conclusion}
\label{sec:conclusion}
A base model can author its own decomposed, state-grounded evaluation suite, freeze it, and use it with no ground truth to gate a keep-best retry loop and to label memory. Under a matched five-repeat protocol on two benchmarks, that system scores above the plain agent without any labels, level with the labelled methods on tau2 but behind ACE on AppWorld; the retry carries the effect, the retry is low-risk because the suites are conservative, and the remaining headroom is selection. The ten labels a subject-matter expert can provide at onboarding add a little more, in either of two ways: as evaluator calibration they lift \method{} by a small margin that was non-negative in every repeat, and as expert-graded warm memory they lift a strong context-engineering learner to tie it. Replacing hand-authored evaluators with self-designed ones is therefore practical with no labels at all and slightly better with the few an expert actually has; seeding the self-designed system's memory from the same onboarding experience, and improving its selector, are the next steps.

{\small
\bibliographystyle{abbrvnat}
\bibliography{references}}

\appendix
\section{Calibration outcomes}
\label{app:cal}
\begin{table}[h]
\centering
\caption{\textbf{Warm-up-v2 calibration outcome per world (campaign A).} Agreement of the authored suite ($S_0$) and of the calibrated proposal with the four gate labels, and the frozen result.}
\footnotesize
\begin{tabular}{l cc l l}
\toprule
World & $S_0$ gate agr. & Proposal gate agr. & Decision & Frozen agg., $\tau$ \\
\midrule
tau2 airline & 100.0\% & 75.0\% & rejected, $S_0$ kept & min, 0.99 \\
tau2 retail & 66.7\% & 83.3\% & \best{accepted} & min, 0.99 \\
tau2 telecom & 100.0\% & 100.0\% & no improvement possible, $S_0$ kept & weighted, 0.85 \\
AppWorld normal & 100.0\% & 100.0\% & no improvement possible, $S_0$ kept & min, 0.99 \\
AppWorld challenge & 100.0\% & 100.0\% & no improvement possible, $S_0$ kept & weighted, 0.85 \\
\bottomrule
\end{tabular}
\end{table}

\section{Authored suites}
\label{app:suites}
The calibrated suites deployed in campaign D contain:
\begin{itemize}
\item airline: two judges, \texttt{final\_state\_outcome\_verifier} and\\ \texttt{policy\_and\_confirmation\_guard}; min, $\tau=0.99$;
\item retail: one judge, \texttt{state\_outcome\_verifier}; min, $0.99$ (calibration dropped a policy-compliance guard);
\item telecom: three judges, \texttt{final\_mms\_outcome\_verifier},\\ \texttt{policy\_and\_precondition\_guard} and \texttt{diagnostic\_coverage\_judge}; weighted, $0.85$;
\item AppWorld normal: two judges, \texttt{final\_state\_outcome\_verifier} and\\ \texttt{unintended\_side\_effect\_guard}; min, $0.99$.
\end{itemize}
Fresh suites in the single-run GPT-5.5 study contained two (retail, telecom, AppWorld normal) or three (airline) judges, with \texttt{min} or \texttt{weighted} aggregation and thresholds of $0.85$ to $0.99$. Every judge is conditioned on the state read-back.

\section{Full matched-campaign table and the label-budget grid}
\label{app:grid}
\begin{table}[h]
\centering
\caption{\textbf{All arms of campaign A (GPT-5.5, five matched repeats).} ``Oracle'' rows read the true label on every task and are not deployable. ``Collapsing'' and ``delta'' are our re-implementation of ACE, with and without a full playbook rewrite after every task; rows labelled plain ``ACE'' use the memory machinery released by the ACE authors. Warm playbooks are fitted from the ten warm-up labels.}
\label{tab:allarms}
\small
\begin{tabular}{l l ccc}
\toprule
Arm & Signal & Tau2 & AppWorld & Overall \\
\midrule
Plain agent & none & $74.4\pm5.4$ & $78.4\pm9.6$ & $76.4\pm6.2$ \\
ACE re-impl.\ collapsing, cold & self-graded & $72.8\pm10.7$ & $90.4\pm3.6$ & $81.6\pm4.6$ \\
ACE re-impl.\ collapsing, cold & calibrated evaluator & $72.8\pm9.6$ & $92.0\pm4.0$ & $82.4\pm3.0$ \\
ACE re-impl.\ collapsing, cold & GT oracle & $72.0\pm4.9$ & $88.8\pm5.2$ & $80.4\pm3.6$ \\
ACE re-impl.\ collapsing, warm, frozen & none & $72.8\pm7.2$ & $88.8\pm5.2$ & $80.8\pm4.6$ \\
ACE re-impl.\ collapsing, warm & self-graded & $69.6\pm8.3$ & $88.0\pm7.5$ & $78.8\pm6.1$ \\
ACE re-impl.\ collapsing, warm & calibrated evaluator & $72.8\pm4.4$ & $92.8\pm5.2$ & $82.8\pm2.3$ \\
ACE re-impl.\ delta, cold & self-graded & $75.2\pm6.6$ & $92.0\pm4.0$ & $83.6\pm4.1$ \\
ACE re-impl.\ delta, cold & calibrated evaluator & $75.2\pm12.5$ & $91.2\pm4.4$ & $83.2\pm4.8$ \\
ACE re-impl.\ delta, cold & GT oracle & $69.6\pm11.2$ & $91.2\pm3.4$ & $80.4\pm6.7$ \\
ACE, cold & self-graded & $71.2\pm8.7$ & $91.2\pm3.4$ & $81.2\pm3.6$ \\
ACE, warm & self-graded & $77.6\pm6.1$ & $90.4\pm3.6$ & $84.0\pm3.2$ \\
ACE, warm & calibrated evaluator & $76.8\pm4.4$ & $92.0\pm4.9$ & $84.4\pm2.6$ \\
Windowed rubric co-evolution (K=20) & evolving suite & $74.4\pm4.6$ & $88.0\pm5.7$ & $81.2\pm3.0$ \\
\method{}, fresh suite & frozen suite & $78.4\pm7.8$ & $86.4\pm2.2$ & $82.4\pm3.9$ \\
\method{} + calibrated suite & frozen suite & $79.2\pm5.9$ & $89.6\pm3.6$ & $84.4\pm4.6$ \\
\bottomrule
\end{tabular}
\end{table}
Three readings of Table~\ref{tab:allarms} inform \S\ref{sec:discussion}. Grading ACE's deployment trajectories with the calibrated \fde{} suite instead of trace self-inference, on top of the warm playbook, changes nothing measurable ($84.4\pm2.6$ versus $84.0\pm3.2$ overall; paired $+0.4\pm2.6$\pp{}, $[-2.8,+3.6]$). Giving ACE the true label on every task does not help ($80.4$ on both memory rows, below the label-free rows): once lessons are discarded by the per-task rewrite, no signal can matter, and disabling the rewrite (delta) is worth $+2.0\pm2.0$\pp{} with the same signal. On the collapsing implementation the calibrated evaluator's role is to prevent drift: warm-then-frozen $72.8$, warm with label-free adaptation $69.6$ ($-3.2\pm3.4$ paired), warm with calibrated-evaluator adaptation $72.8$ on tau2.

\section{Exact values behind Figures~\ref{fig:attempts} and~\ref{fig:headroom}}
\label{app:numbers}
\begin{table}[h]
\centering
\caption{\textbf{Operating point of the calibrated suites (campaign D; mean over five deployments on the whole task sets), values plotted in Figure~\ref{fig:attempts}.} Judges and aggregation as frozen after calibration; per-attempt false-positive (FP) and false-negative (FN) rates of the suite verdict against ground truth; discrimination = mean judge score on GT-pass minus GT-fail attempts; retry gain = deployed minus first-attempt success; oracle = success if the better of the two attempts had always been selected.}
\label{tab:operating}
\footnotesize
\begin{tabular}{l c c c c c c c c c}
\toprule
World & Judges, agg., $\tau$ & Judged pass & GT & Retried & FP & FN & Discr. & Retry gain & Oracle \\
\midrule
Airline (40) & 2, min, 0.99 & 67.5\% & 64.0 & 37.0\% & 39\% & 26\% & $+0.18$ & $+1.5$ & 66.0 \\
Retail (104) & 1, min, 0.99 & 86.5\% & 84.2 & 26.9\% & 53\% & 25\% & $+0.13$ & $+1.0$ & 88.5 \\
Telecom (104) & 3, weighted, 0.85 & 72.3\% & 79.6 & 38.8\% & 13\% & 23\% & $+0.38$ & $+4.6$ & 84.2 \\
AppWorld normal (158) & 2, min, 0.99 & 84.7\% & 90.8 & 20.0\% & 29\% & 23\% & $+0.43$ & $+1.5$ & 91.1 \\
\bottomrule
\end{tabular}
\end{table}

\begin{table}[h]
\centering
\caption{\textbf{Two-model single-run study (campaign C), values plotted in Figure~\ref{fig:headroom}.} Success (\%) on the full suites; one run per cell. \best{Bold} = best in row.}
\label{tab:twomodel}
\small
\begin{tabular}{ll ccc}
\toprule
Model & World & Plain agent & ACE, no GT & \method{} \\
\midrule
\multirow{5}{*}{GPT-5.5 @low} & AppWorld normal (168) & 87.5 & 90.5 & \best{91.7} \\
 & AppWorld challenge (417) & 64.5 & 71.2 & \best{77.7} \\
 & tau2 airline (50) & 64.0 & \best{70.0} & 66.0 \\
 & tau2 retail (114) & \best{83.3} & \best{83.3} & \best{83.3} \\
 & tau2 telecom (114) & 70.2 & 69.3 & \best{75.4} \\
\midrule
\multirow{4}{*}{Opus-4.8 @off} & AppWorld normal (168) & 70.8 & 86.9 & \best{92.3} \\
 & tau2 airline (50) & 68.0 & \best{70.0} & 66.0 \\
 & tau2 retail (114) & 82.5 & \best{85.1} & \best{85.1} \\
 & tau2 telecom (114) & 62.3 & 69.3 & \best{73.7} \\
\bottomrule
\end{tabular}
\end{table}

% \section{Reproducibility}
% \label{app:repro}
% One reported run of the fresh-suite treatment:
% \begin{quote}\footnotesize\ttfamily
% run\_dynamic.py --bench <bench> --dataset <world> --model <model> \textbackslash\\
% \ \ --n-samples 3 --no-briefing --no-revise --require-state-signal \textbackslash\\
% \ \ --balanced-memory --max-attempts 2 --select-margin 0.30 \textbackslash\\
% \ \ --task-ids-file <held-out ids> --rng-seed <seed>
% \end{quote}
% The calibrated treatment adds \texttt{--load-suite <calibrated suite>}, where the suite is produced by\\ \texttt{warmup\_v2\_calibration.py} from the ten reserved warm-up tasks. Campaign supervisors persist every process, seed, suite and progress count, resume a cell only with its own saved suite (a cell never mixes two designed suites) and retry failed executions; result mirrors carry SHA-256 manifests.

\end{document}